\documentclass[10pt,twocolumn,letterpaper]{article}
\usepackage{cvpr}    
\definecolor{cvprblue}{rgb}{0.21,0.49,0.74}
\usepackage[pagebackref,breaklinks,colorlinks,allcolors=cvprblue]{hyperref}
\usepackage{xcolor}
\usepackage{arydshln}
\usepackage{booktabs}
\usepackage{pifont}
\usepackage{float}
\usepackage{algorithm}
\usepackage{algorithmicx}
\usepackage{multirow}
\usepackage{adjustbox}
\usepackage[noend]{algpseudocode}
\usepackage{booktabs}
\usepackage{multirow}
\usepackage{makecell}
\usepackage{threeparttable}
\usepackage[table]{xcolor}
\usepackage{graphicx}

\newcommand{\best}[1]{\textbf{\textcolor{red}{#1}}}
\newcommand{\sbest}[1]{\underline{\textcolor{blue}{#1}}}
\usepackage{makecell}

\newcommand{\valwithx}[2]{\makecell{#1 \\ {\color{gray}(#2$\times$)}}}
\newcommand{\supp}{{\textbf{\textcolor{blue}{Supplementary Material}}}}
\def\@onedot{\ifx\@let@token.\else.\null\fi\xspace}
\def\eg{\emph{e.g}\onedot} 
\def\ie{\emph{i.e}\onedot}

\DeclareUnicodeCharacter{03B1}{\ensuremath{\alpha}}

\def\paperID{2079}
\def\confName{CVPR}
\def\confYear{2026}
\title{
\includegraphics[height=2em]{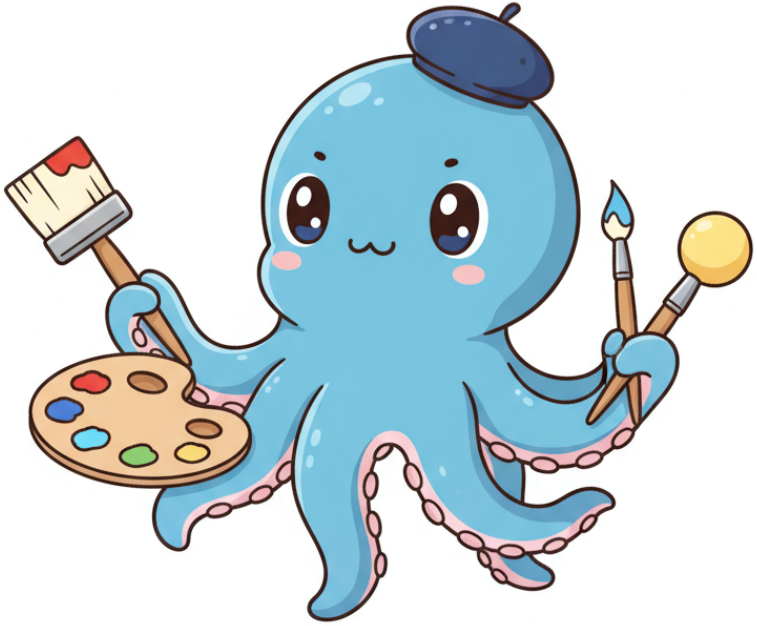}
\hspace{-0.6em}
OctoT2I: A Self-Evolving Agentic Text-to-Image Router}
\author{
    Xu Jiang$^{1,*}$ \quad 
    Bin Chen$^{1,*}$ \quad 
    Gehui Li$^{1,*}$ \quad 
    Yule Duan$^{1,*}$ \quad 
    Ronggang Wang$^{1,2}$ \quad 
    Jian Zhang$^{1,2,\dagger}$ \\
    $^{1}$School of Electronic and Computer Engineering, Peking University \\
    $^{2}$Guangdong Provincial Key Laboratory of Ultra High Definition Immersive Media Technology, \\
    Shenzhen Graduate School, Peking University \\
    {\tt\small \{xjiang25,chenbin,ligehui921,ylduan25\}@stu.pku.edu.cn} 
    {\tt\small \ zhangjian.sz@pku.edu.cn}
}
\begin{document}
\nocite{*}

\twocolumn[{
\renewcommand\twocolumn[1][]{#1}
\maketitle
\centering
\vspace{-0.6cm}
\includegraphics[width=\textwidth]{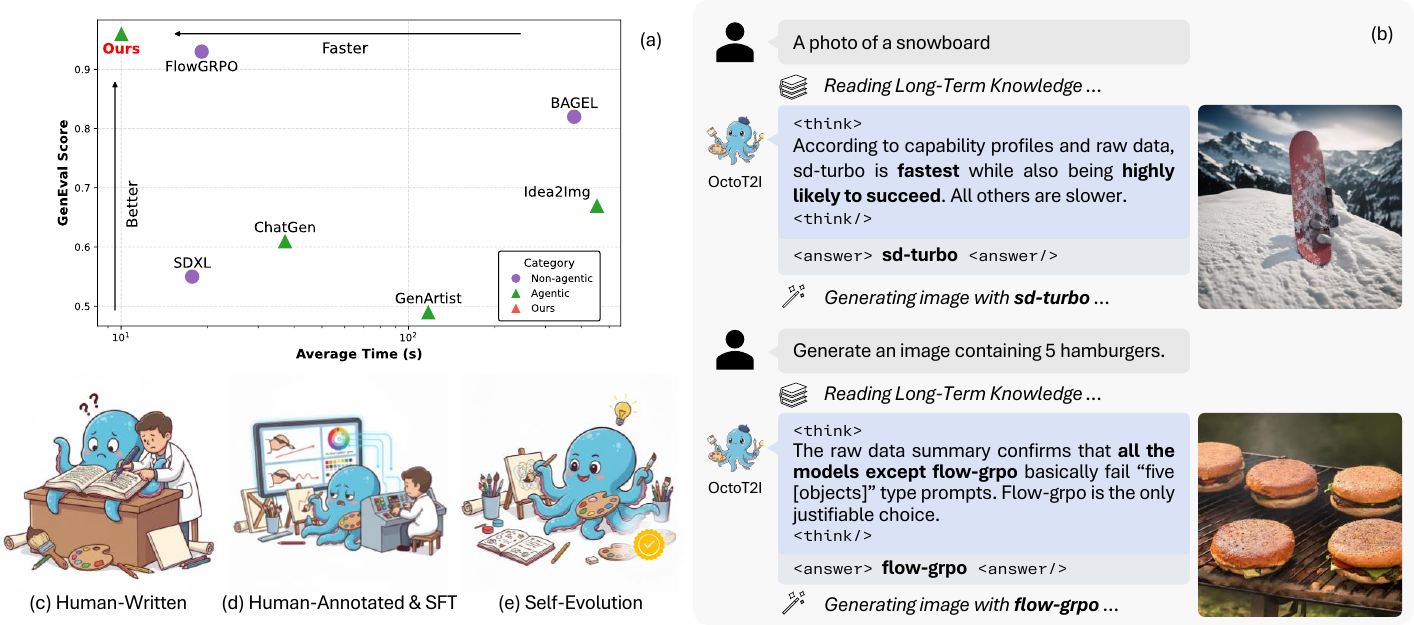}
\vspace{-0.5cm}
\captionsetup{type=figure}
\caption{\textbf{The core advantages of OctoT2I in performance, decision-making, and knowledge acquisition.} 
    \textbf{(a)} OctoT2I (Ours) achieves an exceptional performance-efficiency balance on GenEval. 
    \textbf{(b)} This superior performance stems from its intelligent, evidence-based decisions, which route each user prompt to the most suitable T2I model. For example, it selects the most efficient tool (sd-turbo) for ``a photo of a snowboard'' while allocating the most justifiable tool (flow-grpo) for ``Generate an image containing 5 hamburgers''. 
    \textbf{(c, d, e)} This intelligent decision-making capability is enabled by our novel \textbf{(e)} self-evolving mechanism, which overcomes the limitations of previous \textbf{(c)} handcrafted priors or \textbf{(d)} costly supervised fine-tuning (SFT) with human-annotated data.
    }
    \label{fig:teaser}
}\vspace{0.3cm}]

\maketitle

\def\thefootnote{\fnsymbol{footnote}}
\footnotetext{This work was supported in part by National Natural Science Foundation of China (62372016), National Science and Technology Major Project-Mobile Information Networks (2024ZD130060), Guangdong Provincial Key Laboratory of Ultra High Definition Immersive Media Technology (2024B1212010006), Shenzhen Science and Technology Program (SYSPG20241211173440004) and Outstanding Talents Training Fund in Shenzhen.}
\footnotetext{$^*$Equal contribution. $^\dagger$Corresponding author.}
\def\thefootnote{\arabic{footnote}}

\begin{abstract}
The explosive growth of Text-to-Image (T2I) models, from large-scale versions to lightweight, real-time ones, now faces diminishing marginal returns from single-model scaling. Agentic T2I methods emerged to alleviate this bottleneck by using multiple models. However, existing agentic T2I methods suffer from three key challenges: reliance on expensive handcrafted priors or human annotations, rigid single-path decision mechanisms, and a neglect of inference efficiency. To address these challenges, we introduce \textbf{OctoT2I}, a novel agentic framework that reformulates the T2I task as a joint optimization of generation quality and inference efficiency. OctoT2I implements a stateful, multi-round routing strategy that adaptively selects the most suitable tool based on its knowledge and memory. This strategy is enabled by a knowledge base built from scratch by our novel Self-Evolving Mechanism. 
% Requiring no human supervision, this mechanism employs an iterative ``Propose--Solve--Evaluate--Learn" (PSEL) loop. The PSEL loop autonomously explores combinations of task aspects to discover each tool's capability frontier efficiently, driving continuous self-improvement without external guidance.
This mechanism, which requires no human supervision, first autonomously defines foundational Conceptual Dimensions (\eg, style, color, count) and then intelligently explores their combinations via an iterative ``Propose--Solve--Evaluate--Learn" (PSEL) loop. The PSEL loop efficiently discovers each tool's capability frontier, driving continuous improvement without external guidance.
Extensive experiments demonstrate that OctoT2I achieves competitive performance (\textbf{0.96}) on GenEval while delivering a \textbf{90.3\%} inference speedup and a \textbf{56.6\%} energy-efficiency gain over the leading baseline (Flow-GRPO), striking an exceptional balance between performance and efficiency. Code is available at \url{https://github.com/JaxJiang2642081986/OctoT2I}.

% Extensive experiments demonstrate
%that OctoT2I achieves leading 

% performance (0.96) while delivering an order-of-magnitude inference speedup (e.g., 11.7x–45.2x) and significant energy-efficiency gains over previous agentic methods.
\end{abstract}

\section{Introduction}
\label{sec:intro} 
Text-to-Image (T2I) models translate textual language into corresponding images and thereby improve the efficiency of visual content creation. Architectural progress has been rapid: diffusion models~\cite{chen2026improved,podell2023sdxl,rombach2022high,xie2025sana,zhang2025alignedgen,yang2026gencompositor,chen2025adversarial,sheng2024realosr} now dominate, while autoregressive approaches~\cite{xie2024show, chen2025janus,ma2025janusflow,wang2024emu3} continue to advance. In parallel, the field shows a clear split in scale. One line of work~\cite{esser2024scaling, lin2024sdxl,liu2025flow,wu2025Qwen-image,flux2024} focuses on building larger models to reach state-of-the-art performance, often with reinforcement learning during post-training; for example, reported parameter counts range from roughly \(\sim\)1B in early Stable Diffusion~\cite{rombach2022high} variants and 0.6B in PixArt-\(\alpha\)~\cite{pixart-alpha} to multi-billion scales such as 8B in Stable Diffusion 3~\cite{esser2024scaling} and 24B in Playground v3~\cite{liu2024playground}. A second line of work~\cite{sauer2024adversarial,lin2024sdxl,xie2025sana,chen2025sana} emphasizes efficiency and yields lightweight models with much faster generation. Despite these gains, the marginal return from scaling a single model is diminishing, and the rapidly growing number of different models makes it difficult for non-professional users to select the most suitable tool for a given task.

% \begin{figure}[t]
% \centering
% \includegraphics[width=\linewidth]{Figs/pareto_efficiency_performance.pdf}
% \caption{Pareto efficiency on GenEval. OctoT2I outperforms all compared agentic and non-agentic methods, achieving a leading performance-speed tradeoff.}
% \label{fig:intro_case_example}
% \end{figure}

To overcome the capability bottleneck of a single model, researchers have brought agents~\cite{li2023camel,chen2023autoagents,yang2024gpt4tools,hong2023metagpt,zhao2024diffagent} into T2I. The core idea is to use an LLM as a controller that composes and schedules multiple off-the-shelf T2I models (referred to as ``tools'') so the system can tackle complex tasks that a single tool struggles to solve. Initial works~\cite{wang2024genartist,yang2023idea2img,jia2024chatgen,zhao2024diffagent} provide preliminary evidence that these agentic methods can enhance the capability of T2I systems.

Despite early progress, the performance of agentic T2I methods remains constrained by how they acquire tool-usage knowledge. On the one hand, \cite{wang2024genartist, T2I-Copilot} relies on human-written priors for decision making. Its performance, therefore, depends on the granularity and coverage of human-written tool descriptions, which consequently limit generalization and system performance. On the other hand, \cite{jia2024chatgen,zhao2024diffagent} pursues better generalization by training LLMs on large, finely annotated datasets created by humans. This path is very expensive for both annotation and model training, exhibits limited scalability~\cite{villalobos2024position,sutskever2014sequence}, and is ultimately bounded by human supervision~\cite{hughes2024open, zhao2025absolute}. Beyond knowledge, current systems are further limited by static and single-path decision mechanisms. Workflows such as~\cite{wang2024genartist, jia2024chatgen, T2I-Copilot} typically use only one specific T2I tool for a single round of generation. Idea2Img~\cite{yang2023idea2img} enables multiple rounds of generation, but the generation tool used remains fixed. Finally, efficiency has been overlooked in previous work, leading to high inference latency and computational cost that harm user experience in interactive settings and hinder practical deployment and sustainability in the long term.

To this end, we reformulate agentic T2I as a joint optimization of generation quality and inference efficiency. We then introduce \textbf{OctoT2I}, a novel agentic framework designed to operationalize this new formulation. The core of this framework is a router agent capable of iterative decision-making, which relies on two key information sources: a Knowledge Module for storing long-term knowledge of tools, and a Memory Module for recording the execution history of the current task. In each decision round, the agent queries these two sources and decides which tool to execute. Subsequently, the Evaluation Module quantitatively scores the outcome, and this score is fed back into the memory as new experience to inform the next round of decisions, forming a complete ``reason-act-reflect" loop.

Additionally, to alleviate the performance limitations and high costs in the knowledge acquisition of prior methods, we design a novel \textbf{Self-Evolving Mechanism} where the agent autonomously acquires knowledge from scratch. Specifically, the agent first autonomously defines a set of conceptual dimensions. Subsequently, the agent initiates an iterative ``Propose–Solve–Evaluate–Learn" (PSEL) loop over conceptual dimension combinations, which is guided by an Exploration Space Pruning strategy for efficient and adaptive tool capability frontier search.
In the ``Propose" stage, it instantiates current dimension combination into concrete textual prompts. The agent then proceeds to the ``Solve'' and ``Evaluate" stages by solving tasks with different tools and quantitatively assessing the generated images. Finally, in the ``Learn" stage, it analyzes and summarizes the evaluation results to update its knowledge base. This data-driven process, built entirely on self-interaction, enables the agent to surpass the performance limitations of traditional, handcrafted priors.

\vspace{3pt}
\noindent \ding{113}~(1) We propose OctoT2I, a novel agentic router to operationalize the co-optimization of performance and efficiency. It implements a stateful, multi-round router that leverages knowledge and memory to adaptively select the most suitable tool for each prompt based on real-time feedback.

\vspace{3pt}
\noindent \ding{113}~(2) We introduce a self-evolving mechanism that enables the agent to acquire knowledge entirely through self-interaction, without external data or human supervision.

\vspace{3pt}
\noindent \ding{113} (3) Extensive experiments show that OctoT2I achieves leading performance (\(\mathbf{0.96}\)) while delivering a \(\mathbf{90.3\%}\) inference speedup and a \(\mathbf{56.6\%}\) energy-efficiency gain over the previous state of the art \cite{liu2025flow}, striking an effective balance between performance and efficiency.

\section{Related Work}
\label{sec:formatting}

% 2.1 Text-to-Image Generation：
% 文本到图像（T2I）生成领域的一条核心研究主线，是通过Scaling Up来探索性能上限。这一趋势旨在通过增加模型参数量和训练计算量，提升生成质量、增强对复杂文本指令的理解能力。例如，Stable Diffusion系列模型的发展清晰地体现了这一路径，其参数规模从早期版本持续增长，直至最新的Stable Diffusion 3模型达到80亿。其他大规模模型如FLUX，显著增强了对复杂文本的组合生成能力。此外，Flow-GRPO，Qwen-Image通过强化学习突破pretrained能力上限；总体而言，Scaling Up是提升T2I模型性能的有效手段，但这种性能的提升也带来了巨大的计算成本，包括高昂的推理延迟和能源消耗。
% 第二条主线则是Efficient T2I，其目标是打造推理速度更快、计算成本更低的轻量级模型。例如，SD-Turbo模型，能够将多步采样过程压缩至单步，实现近乎实时的图像生成。SANA使得笔记本的图形处理器上也能实现高分辨率图像生成。但效率的提升往往伴随着一定程度的性能权衡，尤其是在处理复杂的生成任务时。
% 这两条研究路线的并行发展虽带来了繁荣的社区生态，却也因模型能力与成本的认知门槛，导致用户往往难以做出最优选择，造成资源冗余并损害用户体验。 据我们所知，目前尚没有工作来整合这一生态并解决上述问题，这正是我们工作的切入点之一。
\subsection{Text-to-Image Generation}
Recent advances in T2I generation have followed two prominent lines of development. One direction of research~\cite{esser2024scaling,lin2024sdxl,liu2025flow,liu2024playground,flux2024,wu2025Qwen-image} focuses on scaling up model capacity to push the performance frontier. This trend aims to improve image fidelity and the ability to handle complex textual instructions by increasing the number of parameters and the amount of training compute. The evolution of the Stable Diffusion series exemplifies this direction, with model sizes expanding from early variants~\cite{rombach2022high} ($\sim$1B parameters) to the most recent Stable Diffusion 3~\cite{esser2024scaling} (8B). Other large-scale models, such as FLUX~\cite{flux2024}, have further demonstrated strong compositional capabilities. Meanwhile, works like Flow-GRPO~\cite{liu2025flow} and Qwen-Image~\cite{wu2025Qwen-image} incorporate reinforcement learning in post-training stages to surpass the limitations of pretrained objectives. While scaling has proven effective in boosting performance, it also incurs significant computational costs, including high inference latency and substantial energy consumption.

In parallel, another direction of work~\cite{sauer2024adversarial,lin2024sdxl,xie2025sana,chen2025sana} emphasizes efficient T2I models that aim to reduce inference cost and latency. For example, SD-Turbo~\cite{sauer2024adversarial} compresses the multi-step diffusion process into a single step, enabling near real-time generation. SANA~\cite{xie2025sana,chen2025sana} allows high-resolution image synthesis on consumer-grade GPUs. However, these lightweight approaches often trade off generation quality, especially when handling complex prompts.

The coexistence of these two lines has fostered a diverse T2I ecosystem. However, this diversity also raises a practical challenge: general users often lack the expertise to select the most appropriate model for their specific needs. This can result in suboptimal tool selection, unnecessary computational overhead, and diminished user experience. %For example, even for simple prompts such as “a cat”, users may default to heavyweight models like Flow-GRPO, incurring inference times exceeding 20 seconds, whereas lightweight alternatives such as SD-Turbo can produce visually comparable results within 1 second. These observations reflect a deeper issue—the lack of a unified mechanism that bridges high-performance and efficient T2I models. 
To the best of our knowledge, no prior work has attempted to integrate these two lines of work and consider the above challenges, which represents a key motivation for our work.

% 2.2 Agentic Systems：
% 大型语言模型[] 的快速发展，正推动其角色从单纯的语言处理器，向能够进行复杂推理和工具使用的Agentic System转变。以ToolLLM和GPT4Tools 等为代表的开创性工作，证明了LLM能够通过通过调用外部工具，来解决超越其自身知识的复杂任务。

% 最近，Agentic Text to Image generation吸引了越来越多的研究者们的注意，其旨在使用LLM作为T2I Models的控制器，提升图像生成的自主性。[Idea2Img]利用GPT-4V和SDXL实现了迭代式的自动校验和Prompt改写的workflow；[GoT] 利用LLMs生成布局信息，从而增强对元素的控制。[DiffAgent，GenArtist] 实现了能智能选择T2I Models的智能体模型。[ChatGen] 训练了一个能够处理用户聊天风格的输入并自动生成T2I所需所有组件的模型。

% 尽管上述方法已经在工具调度上取得了很大的成功，但性能仍然受限 due to 知识获取、决策机制，并且计算效率较低，阻碍了实际使用。为此，我们引入了自进化机制以驱动Agent自主构建工具知识，突破了对传统人类手工先验所导致的性能限制。除此之外，我们设计了一个能够动态多轮路由的智能体框架，解决过往决策机制导致的性能限制。另外，我们是第一个平衡效率和性能的Agentic T2I方法，更有使用价值。

\begin{figure*}[t]
  \centering
  \includegraphics[width=0.9\linewidth]{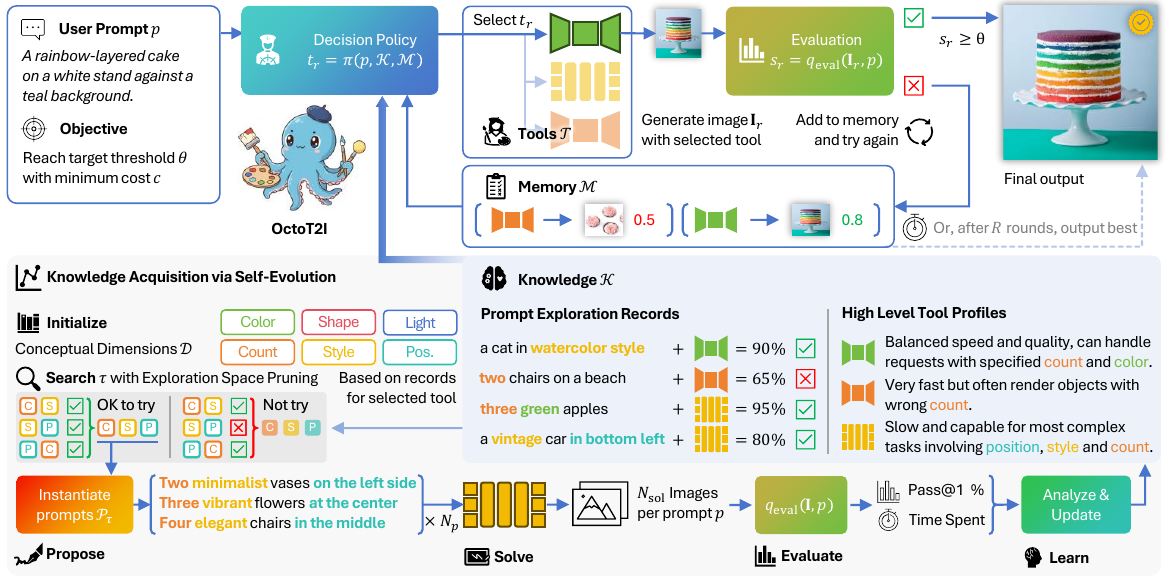}
 \caption{\textbf{The overall architecture of OctoT2I}. 
\textbf{(Top) Inference Workflow.} The agent takes a user prompt and executes a ``reason-act-reflect" loop: The decision policy leverages its memory and knowledge modules to select an appropriate tool. This tool is then used to generate an image, which is subsequently evaluated to obtain a quality score. The workflow terminates when the score meets the quality threshold or the round limit is reached.
\textbf{(Bottom) Self-Evolving Mechanism.} The knowledge module is autonomously and adaptively constructed from scratch. It includes hierarchical high-level tool profiles and exploration records. This construction is driven by the ``Propose–Solve–Evaluate–Learn'' (PSEL) loop, which is guided by the exploration space pruning strategy.}
  \label{fig:method_main}
  \vspace{-10pt}
\end{figure*}

\subsection{Agentic Systems}
The rapid advancement of LLMs~\cite{openai2023gpt,bai2023qwen} has catalyzed their transformation from mere language processors into agentic systems~\cite{wu2023autogen,chen2023autoagents,shen2024hugginggpt,jiang2025multi,hu2026talkphoto,jin2026dgpo,huang2025unishield} capable of controlling how they accomplish tasks and tool usage. Pioneering works such as ToolLLM~\cite{qin2023toolllm} and GPT4Tools~\cite{yang2024gpt4tools} have demonstrated that LLMs can orchestrate external APIs and tools to solve complex tasks that extend beyond their intrinsic knowledge.

Recently, this agentic paradigm has been applied to T2I to overcome the limitations of single models. Early efforts~\cite{wang2024genartist,yang2023idea2img,zhao2024diffagent,jia2024chatgen, T2I-Copilot} in this domain have explored several promising directions. Idea2Img~\cite{yang2023idea2img}leverages an LLM for automated validation and prompt rewriting. DiffAgent~\cite{zhao2024diffagent} and GenArtist~\cite{wang2024genartist} have implemented agents for intelligent T2I model selection, while ChatGen~\cite{jia2024chatgen} trained a model to handle conversational inputs and automatically assemble all components required for generation.

Despite the initial success of existing agentic T2I methods in managing multiple tools, their performance remains constrained due to knowledge acquisition and their static, single-path decision mechanisms. Furthermore, efficiency is overlooked, which hinders practical deployment. To solve these problems, our work introduces a self-evolving mechanism to enable the agent to autonomously construct tool knowledge, thereby surpassing the limitations of handcrafted priors. We also design a dynamic, multi-round routing framework to overcome the constraints of static decision-making. As the novel agentic T2I method to balance performance and efficiency, our approach offers significantly enhanced practical viability.

\section{Methodology}
\label{sec:methodology}

\subsection{Problem Formulation}
\label{sec:problem_formulation}

To operationalize the co-optimization of generation quality and inference efficiency, we define routing as a constrained selection problem over a library of pretrained T2I models. 

\textbf{Setting.} Let $\mathcal{P}$ be the space of prompts and $\mathcal{T}=\{\mathrm{t}_1,\dots,\mathrm{t}_N\}$ a library of tools. Invoking $\mathrm{t}_i$ on a prompt $p\in\mathcal{P}$ returns an image $\mathbf{I}=\mathrm{t}_i(p)$. $\mathrm{q}(\mathbf{I},p)$ denotes the ground-truth generation quality when $\mathrm{t}_i$ is run on prompt $p$, and $\mathrm{c}(\mathrm{t}_i)$ denotes the ground-truth computational cost of $\mathrm{t}_i$.

\textbf{Objective.}
Given a user-acceptable quality threshold $\theta$, the ideal tool $\mathrm{t^\ast}$ is the cost-minimal tool that satisfies the quality requirement:
% \begin{equation}
% \label{eq:problem_definition_oracle}
% \begin{aligned}
% \mathrm{t^\ast(p)} \;=\; \operatorname*{arg\,min}_{\mathrm{t}_i \in \mathcal{T}} \quad & \mathrm{c}(\mathrm{t}_i) \\
% \text{s.t.} \quad & \mathrm{q}(\mathbf{I}, p) \;\ge\; \theta \, .
% \end{aligned}
% \end{equation}
\begin{equation}
\label{eq:problem_definition_oracle}
\mathrm{t^\ast}(p) = \operatorname*{arg\,min}_{\mathrm{t}_i \in \mathcal{T}} \; \mathrm{c}(\mathrm{t}_i) \; \text{s.t.} \; \mathrm{q}(\mathbf{I}, p) \ge \theta \, .
\end{equation}

\textbf{Challenges.}
While Eq.~(\ref{eq:problem_definition_oracle}) defines the ideal objective, the agent lacks prior knowledge of the quality $\mathrm{q}(\cdot)$ and cost $\mathrm{
c}(\cdot)$ functions for each tool. Consequently, we must estimate these quantities from prior data and execute routing decisions robustly in the presence of such estimation error.

\subsection{OctoT2I System Design}
\label{sec:t2i_router_system_design}

\textbf{System Overview.} To solve the problem in Sec.~\ref{sec:problem_formulation}, OctoT2I adopts a multi-round, dynamic routing mechanism. As shown in Fig.~\ref{fig:method_main} (top), at each round $r$, the agent selects a tool $\mathrm{t}_r$ by conditioning on the user prompt $p$, the knowledge $\mathcal{K}$, and the prompt-specific working memory from the previous rounds, $\mathcal{M}_{r-1}$:
\begin{equation}
\label{eq:decision_func}
\mathrm{t}_r \;=\; \mathrm{\pi}(p,\, \mathcal{K},\,\mathcal{M}_{r-1}),
\end{equation}
where $\mathrm{\pi}(\cdot)$ is the decision policy. The agent then executes the chosen tool $\mathrm{t}_r$ on the prompt $p$ to generate an image $\mathbf{I}_r$:
\begin{equation}
\label{eq:action_func}
\mathbf{I}_r =  \mathrm{t}_r(p).
\end{equation}
Then the agent assesses the quality of $\mathbf{I}_r$, by calling the function $\mathrm{q}_\text{eval}$ to yield a score $s_r$:
\begin{equation}
\label{eq:evaluation_func}
s_r \;=\; \mathrm{q}_\text{eval}(\mathbf{I}_r,\, p).
\end{equation}
The outcome of round $r$ (\ie, $\mathrm{t}_r$, $\mathbf{I}_r$, $s_r$) is appended to the memory to form $\mathcal{M}_{r}$, which provides context for the next decision. The workflow terminates when the score meets the quality threshold $\theta$ or when the round limit $R$ is reached, and the highest-scoring image is provided to the user.

\textbf{Knowledge} $\mathcal{K}$ serves as a long-term repository of prior knowledge about the toolset $\mathcal{T}$, providing the foundation for the reasoning process of decision policy $\mathrm{\pi}$. This knowledge covers key performance characteristics such as each tool's core capabilities, applicable scenarios, and inference cost. Crucially, $\mathcal{K}$ is initially empty and is autonomously constructed, which is detailed in Sec.~\ref{sec:self-evolving mechanism}.

\textbf{Memory} $\mathcal{M}$ in round $r$, denoted $\mathcal{M}_r$, is a short-term working memory that is reinitialized for each new prompt. It consists of two components: a sequence of tuples $(\mathrm{t}_j, s_j, \mathbf{I}_j)$ recording the tool, score, and corresponding image for each step up to round $r$, and the best-so-far result $(\mathbf{I}_\text{best}, s_\text{best})$, which tracks the highest score encountered. The memory is defined as follows:
\begin{equation}
\label{eq:memory_trajectory}
\mathcal{M}_{r} = \big(\{(\mathrm{t}_j, s_j, \mathbf{I}_j)\}_{j=1}^{r}, (\mathbf{I}_\text{best}, s_\text{best})\big).
\end{equation}
The best-so-far result is updated after each evaluation as:
\begin{equation}
\label{eq:memory_update}
(\mathbf{I}_\text{best}, s_\text{best}) \leftarrow \begin{cases} (\mathbf{I}_r, s_r) & \text{if } s_r > s_\text{best} \\ (\mathbf{I}_\text{best}, s_\text{best}) & \text{otherwise} \end{cases}.
\end{equation}
% \begin{equation}
% (\mathbf{I}_\text{best}, s_\text{best}) = \left( \mathbf{I}_k, \max \{s_j\}_{j=1}^{r} \right) \quad \text{where} \quad k = \operatorname*{argmax}_{j \in \{1, \dots, r\}} s_j
% \end{equation}
This memory, capturing both the execution history and the current optimum, provides the decision policy with the necessary information for adaptive routing.

\begin{figure*}[!t]
\centering
\includegraphics[width=0.9\linewidth]{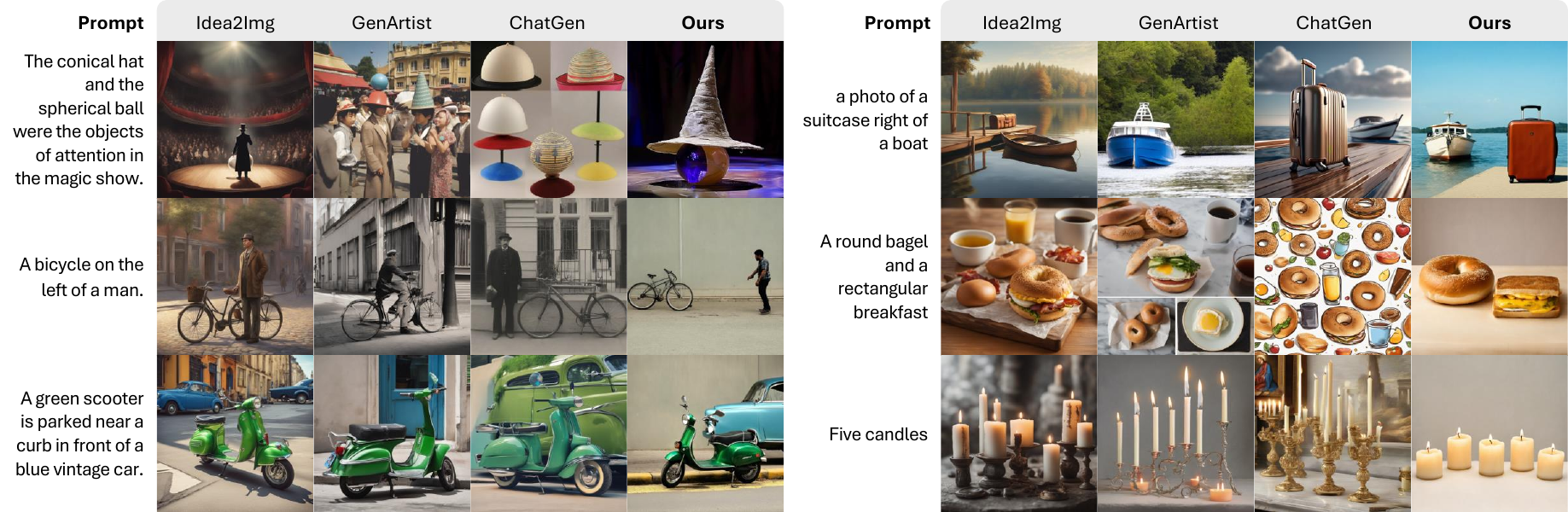}
\caption{\textbf{Qualitative comparison} with agentic T2I methods.}
\label{fig:comparison_qualitative}
% \vspace{-5pt}
\end{figure*}

% \textbf{Decision Policy} $\mathrm{\pi}$ aims to solve the constrained optimization problem in Sec.~\ref{sec:problem_formulation}. This problem is equivalently transformed into a more operational filter-then-select process. At round $r$, the decision policy $\mathrm{\pi}$ first identifies a feasible set comprising all tools whose estimated quality meets the threshold based on the current state (\ie, $p, \mathcal{K}, \mathcal{M}_{r-1}$). Second, the tool with the minimum estimated cost $\hat{\mathrm{c}}$ is selected from this feasible set:
% \begin{equation}
% \label{eq:decision_policy_definition}
% \mathrm{t}_r = \mathrm{\pi}(p, \mathcal{K}, \mathcal{M}_{r-1}) = \operatorname*{arg\,min}_{\mathrm{t}_i \in \mathcal{T}, \hat{\mathrm{q}}({\mathrm{t}_i}(p), p) \ge \theta} \hat{\mathrm{c}}(\mathrm{t}_i).
% \end{equation}
% This policy is implemented via prompting an LLM with an explicit chain-of-thought prompt template $\mathbf{p}_{\text{decision}}$, guiding it to approximate the reasoning in the above process.
\textbf{Decision Policy} $\mathrm{\pi}$ aims to solve the constrained optimization problem in Sec.~\ref{sec:problem_formulation}. We transform this problem into an operational "filter-then-select" process. This policy is implemented by an LLM conditioned on an explicit chain-of-thought (CoT) template, $\mathbf{p}_{\text{decision}}$. At round $r$, this CoT prompt guides the LLM to approximate the reasoning based on the current state (\ie, $p, \mathcal{K}, \mathcal{M}_{r-1}$). First, the LLM queries $\mathcal{K}$ and $\mathcal{M}_{r-1}$ to estimate the quality $\hat{\mathrm{q}}$ for each tool $\mathrm{t}_i$, identifying a feasible set of tools whose estimated quality meets the threshold (\ie, $\hat{\mathrm{q}}({\mathrm{t}_i}(p), p) \ge \theta$). Second, the LLM retrieves the estimated cost $\hat{\mathrm{c}}$ from $\mathcal{K}$ for each valid tool and selects the one with the minimum estimated cost. This LLM-driven reasoning process can be expressed as:
\begin{equation}
\label{eq:decision_policy_definition}
\mathrm{t}_r =\mathrm{\pi}(p, \mathcal{K}, \mathcal{M}_{r-1}) = \operatorname*{arg\,min}_{\mathrm{t}_i \in \mathcal{T}, \hat{\mathrm{q}}({\mathrm{t}_i}(p), p) \ge \theta} \hat{\mathrm{c}}(\mathrm{t}_i).
\end{equation}

% \textbf{Decision Policy} $\mathrm{\pi}$ aims to solve the constrained optimization problem in Sec.~\ref{sec:problem_formulation}. This problem is equivalently transformed into a more operational filter-then-select process. At round $r$, the decision policy $\pi$ first identifies a feasible set $\mathcal{F}(x_r)$ comprising all tools whose estimated quality meets the threshold based on the current state $x_r = (p, K, M_{r-1})$. Second, the tool with the minimum estimated cost $\hat{c}$ is selected from this feasible set:
% \begin{equation}
% \label{eq:decision_policy_definition}
% t_r = \pi(x_r) = \operatorname*{arg\,min}_{t_i \in \mathcal{F}(x_r)} \hat{c}(t_i, x_r).
% \end{equation}
% This policy is implemented via prompting a LLM with an explicit chain-of-thought prompt template $\mathbf{p}_{\text{decision}}$, guiding it to approximate the reasoning in the above process.

\begin{table*}[!t]
\centering
\caption{\textbf{Quantitative evaluation on GenEval benchmark.} Throughout this paper, the best and second-best results are marked in \best{bold red} and \sbest{underlined blue}, respectively. $\uparrow$ indicates higher is better. Obj.: Object; Attr.: Attribution.}
\vspace{-5pt}
\label{tab:GE}
\begin{threeparttable}
\setlength{\tabcolsep}{3pt}
\renewcommand{\arraystretch}{1.1}
\resizebox{0.7\textwidth}{!}{%
\begin{tabular}{l|l|cccccc|c}
\toprule
\textbf{Type} & \textbf{Method} & \textbf{Single Obj.} & \textbf{Two Obj.} & \textbf{Counting} & \textbf{Colors} & \textbf{Position} & \textbf{Color Attri.} & \textbf{Overall}$\uparrow$ \\
\midrule
\multirow{8}{*}{\hspace*{2mm}\rotatebox[origin=c]{90}{\textit{Non-agentic}}}
 % & sd-turbo & \sbest{0.99} & 0.49 & 0.35 & 0.87 & 0.08 & 0.13 & 0.48 \\
 & SDXL-Lightning & 0.98 & 0.69 & 0.36 & 0.85 & 0.09 & 0.18 & 0.52 \\
 % & SDXL & 0.98 & 0.74 & 0.39 & 0.85 & 0.15 & 0.23 & 0.55 \\
 & SDXL-Turbo & \sbest{0.99} & 0.71 & 0.51 & 0.83 & 0.07 & 0.20 & 0.55 \\
 & Show-o & 0.95 & 0.52 & 0.49 & 0.82 & 0.11 & 0.28 & 0.53 \\
 & JanusFlow & 0.97 & 0.59 & 0.45 & 0.83 & 0.53 & 0.42 & 0.63 \\
 & Janus-Pro & \sbest{0.99} & 0.89 & 0.59 & 0.90 & 0.79 & 0.66 & 0.80 \\
 & BAGEL & \sbest{0.99} & \sbest{0.94} & 0.81 & 0.88 & 0.64 & \sbest{0.63} & 0.82 \\
 & FLUX.1-dev & \sbest{0.99} & 0.85 & 0.74 & 0.79 & 0.21 & 0.48 & 0.68 \\
 % & SD3.5-M & 0.98 & 0.78 & 0.50 & 0.81 & 0.24 & 0.52 & 0.63 \\
 % & SANA-1.5 4.8B & \sbest{0.99} & \sbest{0.95} & 0.77 & 0.88 & 0.63 & 0.60 & 0.80 \\
 & Flow-GRPO & \sbest{0.99} & \best{0.99} & \sbest{0.88} & \best{0.95} & \sbest{0.95} & \best{0.86} & \sbest{0.93} \\
\midrule
\multirow{4}{*}{\hspace*{2mm}\rotatebox[origin=c]{90}{\textit{Agentic}}}
 & Idea2Img & 0.98 & 0.81 & 0.68 & 0.93 & 0.25 & 0.41 & 0.67 \\
 % & GoT & 0.98 & 0.70 & 0.63 & 0.80 & 0.29 & 0.25 & 0.61 \\
 & GenArtist & 0.89 & 0.42 & 0.47 & 0.73 & 0.29 & 0.13 & 0.49 \\
 & ChatGen & 0.93 & 0.49 & 0.23 & 0.75 & 0.12 & 0.14 & 0.44 \\
 % \rowcolor{blue!10}
 & \cellcolor{blue!10}\textbf{Ours} & \cellcolor{blue!10}\best{1.00} & \cellcolor{blue!10}\best{0.99} & \cellcolor{blue!10}\best{0.95} & \cellcolor{blue!10}\sbest{0.94} & \cellcolor{blue!10}\best{1.00} & \cellcolor{blue!10}\best{0.86} & \cellcolor{blue!10}\best{0.96} \\
\bottomrule
\end{tabular}}
\end{threeparttable}
\vspace{-10pt}
\end{table*}

\textbf{Evaluation.}
As shown in Fig.~\ref{fig:method_main} (top), the evaluation function $\mathrm{q}_\text{eval}$ provides a quantitative feedback score $s_r$ on the quality of the generated image $\mathbf{I}_r$ given the prompt $p$, inspired by VQA-Score~\cite{vqascore}. Specifically, we employ an MLLM to answer a binary ``yes/no'' question regarding prompt–image alignment. Instead of relying on the discrete textual response~\cite{xie2025sana}, we extract the raw logits for the ``yes'' and ``no'' tokens and softmax them into a continuous score $s_r$. This design yields a fine-grained feedback signal, providing the basis for both adaptive decision-making and the ``Evaluate" stage of self-evolving mechanism in Sec.~\ref{sec:self-evolving mechanism}.

\subsection{Knowledge Acquisition via Self-Evolution}
\label{sec:self-evolving mechanism}

As motivated previously, the major challenge in enabling the policy of Eq.~(\ref{eq:decision_policy_definition}) is how to acquire reliable tool knowledge~$\mathcal{K}$. Rather than relying on handcrafted priors or costly human annotations, we introduce a novel \textbf{self-evolving mechanism} where the agent autonomously and adaptively constructs knowledge $\mathcal{K}$ from scratch.

\textbf{Overview.} 
As shown in Fig.~\ref{fig:method_main} (bottom), the agent first initializes the exploration space by defining a foundational set of conceptual dimensions~$\mathcal{D}$. Subsequently, for each tool $\mathrm{t}_i \in \mathcal{T}$, the agent initiates an iterative PSEL loop. Within this loop, the agent's exploration is guided by the exploration space pruning strategy, which leverages learned knowledge to decide which dimension combinations to explore next. By repeatedly executing this PSEL loop for each tool $\mathrm{t}_i$, the agent progressively constructs the knowledge base $\mathcal{K}_i$.

% \begin{figure}[!t]
% \vspace{-5pt}
% \centering
% \includegraphics[width=0.9\linewidth]{Figs/pareto_efficiency_performance.pdf}
% \vspace{-10pt}
% \caption{Pareto efficiency on GenEval. OctoT2I outperforms all compared agentic and non-agentic methods, achieving a leading performance-speed tradeoff.}
% \label{fig:pareto_efficiency_performance}
% \vspace{-10pt}
% \end{figure}

\textbf{Initialization.} 
The self-evolution mechanism begins by establishing a structured basis for the vast and unstructured prompt space $\mathcal{P}$ to enable systematic exploration. 
Inspired by~\cite{huang2025t2icompbench++,wei2025tiifbenchdoest2imodel,ghosh2023geneval}, the LLM is conditioned on the prompt $\mathbf{p}_{\text{define}}$ to generate $N_\mathcal{D}$ fundamental and orthogonal conceptual dimensions (\eg, color, position, culture and counting), denoted as $\mathcal{D}$. 
These conceptual dimensions serve as building blocks to construct the complete exploration space, defined as the power set of $\mathcal{D}$ excluding the empty set:
\begin{equation}
\label{eq:concept_combinations}
\mathcal{C}_{\text{explore}} = 2^\mathcal{D} \setminus \{\emptyset\}.
\end{equation}
The mechanism explores $\mathcal{C}_{\text{explore}}$ in a simple-to-complex order based on the number of conceptual dimensions in each combination $\tau$, \ie, from $|\tau|=1$ up to $N_\mathcal{D}$.

\textbf{Propose.}  
Given a dimension combination $\tau \in \mathcal{C}_{\text{explore}}$, the agent acts as a task proposer and uses the prompt template $\mathbf{p}_{\text{propose}}$ to instantiate $\tau$ into $N_p$ diverse and concrete textual prompts, denoted as $\mathcal{P}_\tau$. To ensure diversity, the agent maintains a history of previously proposed prompts and generates only new prompts.

\begin{table*}[!t]
\centering
\caption{\textbf{Quantitative evaluation results on T2ICompBench++.} $\uparrow$ indicates higher is better.}
\vspace{-5pt}
\label{tab:++}
\begin{threeparttable}
\setlength{\tabcolsep}{6pt}
\renewcommand{\arraystretch}{1.1}
\begin{adjustbox}{max width=0.85\linewidth}
\begin{tabular}{l|cccccccc|c}
\toprule
\textbf{Method} & \textbf{Color} & \textbf{Shape} & \textbf{Texture} & \textbf{2D-Spatial} & \textbf{3D-Spatial} & \textbf{Numeracy} & \textbf{Non-Spatial} & \textbf{Complex} & \textbf{Average}$\uparrow$ \\
\midrule
\multicolumn{10}{c}{\textit{Non-agentic Methods}} \\
\midrule
SD1.5          & 0.3591 & 0.3720 & 0.3961 & 0.0990 & 0.3139 & 0.4391 & 0.7866 & 0.3121 & 0.3847 \\
SD-Turbo       & 0.5802 & 0.4485 & 0.5415 & 0.1345 & 0.3334 & 0.4979 & 0.8333 & 0.3506 & 0.4650 \\
SANA Sprint & 0.7400 & 0.4987 & 0.6278 & 0.3548 & 0.4050 & 0.5855 & 0.8867 & 0.3734 & 0.5590 \\
SDXL           & 0.6111 & 0.4856 & 0.5201 & 0.1908 & 0.3558 & 0.5089 & 0.9067 & 0.3309 & 0.4887 \\
SD3.5-M        & 0.7119 & 0.5339 & \sbest{0.7262} & 0.2868 & 0.3554 & 0.5700 & 0.8733 & 0.3418 & 0.5499 \\
DALLE 3  & 0.7785  & \sbest{0.6205 } & 0.7036  & 0.2865 & 0.3744  & 0.5296  & 0.9170 & 0.3771 & 0.5734 \\
SANA-1.5  & 0.7735 & 0.5571 & 0.6744 & 0.3706 & 0.4100 & 0.6045 & \sbest{0.9233} & \sbest{0.3856} & 0.5874 \\
Flow-GRPO       & \sbest{0.8119} & 0.6143 & 0.7194 & \sbest{0.5199} & \best{0.4373} & \sbest{0.6752} & 0.9133 & 0.3741 & \sbest{0.6332} \\
\midrule
\multicolumn{10}{c}{\textit{Agentic Methods}} \\
\midrule
Idea2Img       & 0.6366 & 0.4596 & 0.5522 & 0.2799 & 0.3538 & 0.5240 & 0.8933 & 0.3490 & 0.5060 \\
% GoT            & 0.4736 & 0.3271 & 0.4270 & 0.2462 & 0.3788 & 0.4954 & 0.7933 & 0.3223 & 0.4330 \\
GenArtist      & 0.6089 & 0.4797 & 0.5241 & 0.1920 & 0.3357 & 0.5026 & 0.8733 & 0.3291 & 0.4807 \\
ChatGen        & 0.4791 & 0.3628 & 0.3760 & 0.1562 & 0.2963 & 0.4374 & 0.7400 & 0.3157 & 0.3954 \\
\rowcolor{blue!10}
\textbf{Ours}  & \best{0.8344} & \best{0.6270} & \best{0.7319} & \best{0.5272} & \sbest{0.4334} & \best{0.7508} & \best{0.9600} & \best{0.4300} & \best{0.6618} \\
\bottomrule
\end{tabular}
\end{adjustbox}
\end{threeparttable}
\end{table*}

\begin{table}[t]
\centering
\vspace{-5pt}
\caption{\textbf{Efficiency comparison with other competitive methods} on GenEval. $\downarrow$ indicates lower is better.}
\vspace{-5pt}
\label{tab:efficiency}
\setlength{\tabcolsep}{2pt}
\renewcommand{\arraystretch}{1.3}
\scalebox{0.7}{
\begin{tabular}{l|cc|cccc}
\toprule
\textbf{Metrics} &
% \textbf{SDXL} &
\textbf{BAGEL} & \textbf{Flow-GRPO} & \textbf{Idea2Img} 
% & \textbf{GoT}
& \textbf{ChatGen} 
& \textbf{GenArtist} & \cellcolor{blue!10}\textbf{Ours} \\
\hline\hline
Avg. Time (s)$\downarrow$ 
% & \valwithx{\sbest{17.67}}{1.76} 
& \valwithx{377.39}{37.7} 
& \valwithx{19.07}{1.90} 
& \valwithx{453.22}{45.2} 
& \valwithx{37.20}{3.71} 
% & \valwithx{20.22}{2.02} 
& \valwithx{117.29}{11.7} 
& \cellcolor{blue!10}\valwithx{\best{10.02}}{1.00} \\
CO$_2$e (g)$\downarrow$      
% & \valwithx{\sbest{611.06}}{1.09} 
& \valwithx{11198.25}{20.0} 
& \valwithx{878.72}{1.57} 
& \valwithx{12033.28}{21.5} 
& \valwithx{1258.47}{2.25} 
% & \valwithx{783.60}{1.40} 
& \valwithx{4403.95}{7.87} 
& \cellcolor{blue!10}\valwithx{\best{559.50}}{1.00} \\
kWh $\cdot$ PUE$\downarrow$  
% & \valwithx{\sbest{1.41}}{1.09} 
& \valwithx{25.86}{20.0} 
& \valwithx{2.02}{1.57} 
& \valwithx{27.79}{21.6} 
& \valwithx{2.91}{2.26} 
% & \valwithx{1.81}{1.40} 
& \valwithx{10.17}{7.89} 
& \cellcolor{blue!10}\valwithx{\best{1.29}}{1.00} \\
\bottomrule
\end{tabular}
}
% \vspace{-5pt}
\end{table}

\textbf{Solve.}  
In the ``Solve'' stage, each prompt $p_{\tau} \in \mathcal{P}_\tau$ is executed with the current tool $\mathrm{t}_i$. 
To ensure robustness, the agent performs $N_\text{sol}$ independent runs per prompt, producing a set of candidate images $\{\mathbf{I}_{\tau,n}\}_{n=1}^{N_\text{sol}}$ for each $p_{\tau}$.

\textbf{Evaluate.}  
In the ``Evaluation'' stage, the agent assesses the quality of the generated candidates $\{\mathbf{I}_{\tau,n}\}_{n=1}^{N_\text{sol}}$ for each exploration prompt $p_{\tau} \in \mathcal{P}_\tau$. 
For every image $I_{\tau,n}$, the evaluation function $\mathrm{q}_\text{eval}$ in Eq.~(\ref{eq:evaluation_func}) produces an alignment score $s_{\tau,n}$ with respect to $p_{\tau}$. 
Considering the stochasticity of generation and our constrained-optimization objective in Eq.~(\ref{eq:problem_definition_oracle}), we adopt the Pass@1 to estimate the probability that a tool’s single-attempt output meets the quality threshold $\theta$:  
\begin{equation}
\label{eq:pass_at_1}
\mathrm{Pass@1}(p_{\tau}, \mathrm{t}_i) = \frac{1}{N_\text{sol}} \sum_{n=1}^{N_\text{sol}} \mathbb{I}(s_{\tau,n} \ge \theta).
\end{equation}
These evaluation results provide empirical evidence for the subsequent ``Learn" stage, where the knowledge $\mathcal{K}$ is incrementally updated.

\textbf{Learn.}  
In the ``Learn'' stage, as shown in Fig.~\ref{fig:method_main} (bottom), the agent structures the raw outcomes from the ``Solve'' and ``Evaluate'' stages into long-term knowledge along two complementary dimensions.  

First, it consolidates fine-grained empirical evidence by recording explored prompts together with their Pass@1 success rates. These \textbf{prompt exploration records} provide data-driven, bottom-up references that the decision policy can reuse when encountering similar cases during inference.  

Second, the agent abstracts these records into \textbf{high level tool profiles}, which summarize both the average inference cost and a semantic description of the tool’s strengths and limitations. This high-level knowledge offers generalizable, top-down guidance for reasoning about novel prompts without direct precedents.  

This multi-layered knowledge provides the decision policy with a comprehensive and complementary foundation. Additionally, it acts as the basis for more efficient exploration, detailed in the subsequent part.

\begin{table}[!t]
% \vspace{-5pt}
\centering
\setlength{\tabcolsep}{2pt}
\caption{\textbf{Ablation on self-evolving mechanism on GenEval.} 
% We compare our approach with hand-crafted priors and GPT5's internal knowledge.
}
\label{tab:abla_self_evolving}
\vspace{-5pt}
\resizebox{\linewidth}{!}{
\begin{tabular}{l|ccc|c}
\toprule
\textbf{Method} & \textbf{Counting} & \textbf{Position} & \textbf{Color Attri.} & \textbf{Overall$\uparrow$} \\
\hline\hline
GPT Internal Knowledge  & 0.85 & 0.73 & 0.61 & 0.85 \\
Hand-Crafted Prior  & 0.90 & 0.95 & 0.85 & 0.93 \\
\rowcolor{blue!10}
\textbf{Self-Evolving Knowledge (Ours)}  & \best{0.95} & \best{1.00} & \best{0.86} & \best{0.96} \\
\bottomrule
\end{tabular}}
\vspace{-5pt}
\end{table}

% \begin{figure*}[!t]
% \vspace{-20pt}
% \centering
% \centerline{\includegraphics[width=\linewidth]{Figs/Ablation_Knowledgepdf.pdf}}
% \vspace{-5pt}
% \caption{\textbf{Ablation study of the self-evolving mechanism.} The reasoning processes of the decision policy in the first round under (a) GPT Internal Knowledge, (b) Hand-Crafted Prior, and (c) Self-Evolving Knowledge (Ours) are compared.}
% \label{fig:abla_knowledge}
% \vspace{-5pt}
% \end{figure*}

\textbf{Exploration Space Pruning.}
The vast size of the potential exploration space, $C_{\text{explore}}$, makes a brute-force search computationally infeasible. Moreover, our goal is not to serve as a static benchmark for comprehensive performance reporting, but rather as a dynamic explorer that efficiently discovers each tool's unique capability frontier. 

To this end, our exploration space pruning strategy defines which tasks are worth exploring. As illustrated in Fig.~\ref{fig:method_main} (bottom), it follows a recursive prerequisite principle: a complex dimension combination $\tau$ is only considered for exploration only if the agent, based on its existing knowledge, determines that the tool $\mathrm{t}_i$ has already mastered all simpler subtasks composing it ($\forall \tau' \subset \tau, \tau' \neq \emptyset$). This determination of whether a tool has mastered a given subtask is made by checking if the tool's historical average Pass@1 score on prompts proposed for $\tau'$ exceeds the threshold $\theta$. 

Through this mechanism, the agent dynamically constructs a personalized and dynamic exploration plan for each tool. The recursive checking process ensures that computational resources remain focused on tasks located at the capability boundary, thereby enabling the overall self-evolution to be both efficient and effective.

\begin{table}[!t]
\centering
\setlength{\tabcolsep}{4pt}
\caption{\textbf{Ablation on decision policy (DP).}}
\label{tab:abla_router}
\vspace{-5pt}
\resizebox{\linewidth}{!}{
\begin{tabular}{l|ccccc|c}
\toprule
\textbf{Method} & \textbf{Color} & \textbf{Shape} & \textbf{Texture} & \textbf{2D-Spatial} & \textbf{Numeracy}  & \textbf{Average}$\uparrow$ \\
\hline\hline
w/o DP  & 0.7016 & 0.5109 & 0.6342 & 0.3078 & 0.4815 & 0.5379 \\
\rowcolor{blue!10}
\textbf{w/ DP (Ours)} & \best{0.8344} & \best{0.6270} & \best{0.7319} & \best{0.5272} & \best{0.7508} & \best{0.6618} \\
\bottomrule
\end{tabular}}
% \vspace{-5pt}
\end{table}

\section{Experiment}
\subsection{Experimental Setting}
\textbf{Implementation Details.}
Toolset comprises five T2I models: Flow-GRPO~\cite{liu2025flow}, SDXL-Turbo~\cite{sauer2024adversarial}, SD-Turbo~\cite{sauer2024adversarial}, SANA1.5~\cite{xie2025sana}, and SANA-Sprint~\cite{chen2025sana}. We use NVILA-Lite-2B-Verifier~\cite{liu2024nvila} for evaluation module on the GenEval~\cite{ghosh2023geneval} and T2I-CompBench++~\cite{huang2025t2icompbench++} and GPT-4o~\cite{Gpt-4o} on WISE~\cite{niu2025wise}. The core controller of the agent is Qwen2-0.5B~\cite{wang2024qwen2}, obtained via policy distillation from GPT-4.1~\cite{openai2023gpt}. $R$ is 4, $\theta$ is 0.8, $N_\mathcal{D}$ is 7, $N_p$ is 10, and $N_\text{sol}$ is 5. All experiments are conducted on four NVIDIA RTX 3090 GPUs. More details on OctoT2I's workflow, function implementation, policy distillation and prompt templates are provided in the \supp.

\textbf{Compared methods} include SDXL-Lightning~\cite{lin2024sdxl}, SDXL-Turbo~\cite{sauer2024adversarial}, Show-o~\cite{xie2024show}, JanusFlow~\cite{ma2025janusflow}, Janus-Pro~\cite{chen2025janus}, BAGEL~\cite{deng2025bagel}, FLUX.1-dev~\cite{flux2024}, SD1.5~\cite{rombach2022high}, SD-Turbo~\cite{sauer2024adversarial}, SANA Sprint~\cite{chen2025sana}, SDXL~\cite{podell2023sdxl}, SD3.5-M~\cite{esser2024scaling}, DALLE 3~\cite{betker2023improving}, SANA-1.5~\cite{xie2025sana}, Flow-GRPO~\cite{liu2025flow}, Idea2Img~\cite{yang2023idea2img}, GenArtist~\cite{wang2024genartist} and ChatGen~\cite{jia2024chatgen}. All compared methods use official default settings.

\textbf{Evaluation Setups.}
Experiments are conducted on three widely-used benchmarks: GenEval~\cite{ghosh2023geneval}, T2I-CompBench++~\cite{huang2025t2icompbench++} and WISE~\cite{niu2025wise}, as well as on a user study with 30 random prompts from DiffusionDB~\cite{wangDiffusionDBLargescalePrompt2022}. We follow the default evaluation metrics for all benchmarks. Metrics reported as most aligned with human evaluations for T2I-CompBench++ are adopted. The calculation of CO\textsubscript{2}e and kWh$\cdot$PUE metrics follows~\cite{strubell2020energy}.

\begin{table}[!t]
\centering
\setlength{\tabcolsep}{3pt} 
\caption{\textbf{Ablation on exploration space pruning (ESP) strategy.} 
}
\label{tab:abla_pruning}
\vspace{-5pt}
\resizebox{0.9\linewidth}{!}{ 
\begin{tabular}{l|c|cc}
\toprule
\textbf{Method} & \textbf{Overall Score}$\uparrow$ & \textbf{Explored Prompts}$\downarrow$ & \textbf{Avg. Time (s)}$\downarrow$ \\
\hline\hline
w/o ESP & \best{0.96} & 1270 & 6857.4 \\
\rowcolor{blue!10}
\textbf{w/ ESP (Ours)} & \best{0.96} & \best{370} & \best{2328.7} \\
\bottomrule
\end{tabular}
}
\vspace{-15pt}
\end{table}

\subsection{Comparison with Other Methods}
\textbf{Quantitative Comparison.}
Tab.~\ref{tab:GE} and \ref{tab:++} demonstrate that our approach not only surpasses the strongest non-agentic method but also significantly outperforms all previous agentic frameworks. On GenEval, our overall score of 0.96 is substantially higher than that of other agentic methods. This advantage is further validated on T2ICompBench++, where our average score of 0.6618 leads all previous methods. 
% The superiority of OctoT2I across both benchmarks clearly demonstrates that our dynamic, multi-round agentic router with a self-evolved knowledge base, effectively overcomes the core limitations of previous knowledge acquisition and static, single-path decision-making.

\textbf{Efficiency Comparison}
OctoT2I delivers substantial improvements in inference efficiency, outperforming all compared agentic methods across all metrics as shown in Tab.~\ref{tab:efficiency}. Our average inference time achieves a significant speedup over prior agentic methods (\eg, approx. 45x faster than Idea2Img and 11x faster than GenArtist), which is critical for interactive use. Furthermore, our approach is the most computationally and environmentally efficient, recording the lowest carbon emissions (CO\textsubscript{2}e) and energy consumption (kWh$\cdot$PUE), leading to a more sustainable and environmentally responsible real-world deployment.

\textbf{Qualitative Comparison.} Fig.~\ref{fig:comparison_qualitative} visually demonstrates that OctoT2I, benefiting from the knowledge built by self-evolution, accurately analyzes prompts and routes them to the most suitable tool. For instance, when handling challenging spatial relationship prompts (\eg, ``A bicycle on the left of a man''), OctoT2I successfully generates the correct spatial layout. In contrast, Idea2Img, GenArtist, and ChatGen all fail to adhere to the spatial instruction. Furthermore, when faced with prompts requiring the combination of multiple unique objects (\eg, ``The conical hat and the spherical ball...''), OctoT2I correctly generates aesthetic images. GenArtist incorrectly merges the objects with a portrait, while ChatGen generates completely irrelevant objects.

% \begin{figure}[!t]
% % \vspace{-20pt}
% \centering
% \includegraphics[width=0.8\linewidth]{Figs/ablation_study_max_tries.pdf}
% \vspace{-5pt}
% \caption{\textbf{Ablation study of the maximum number of rounds} on GenEval. Score (green, left) and inference time (purple, right).}
% \label{fig:ablation_max_tries}
% % \vspace{-15pt}
% \end{figure}

\begin{figure}[!t]
\vspace{-10pt}
\centering
\includegraphics[width=0.8\linewidth]{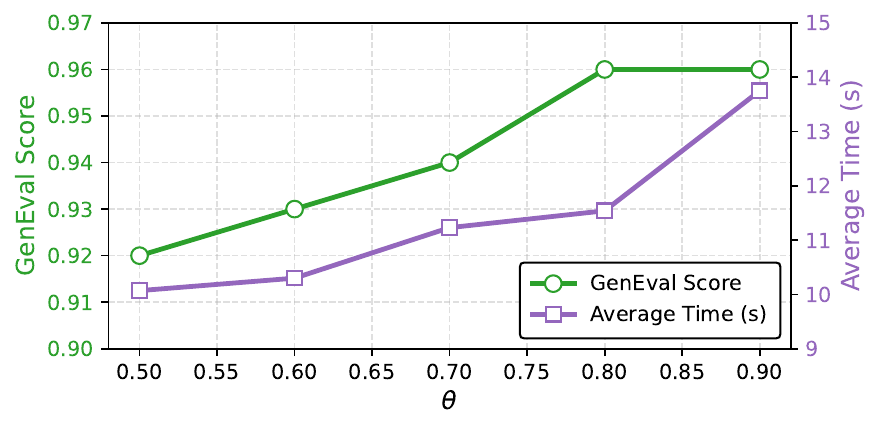}
\vspace{-10pt}
\caption{\textbf{Ablation study of $\theta$} on performance (green, left axis) and inference time (purple, right axis) on the GenEval benchmark.}
\vspace{-10pt}
\label{fig:ablation_threshold}
\end{figure}

\subsection{Ablation Study}
\textbf{Effect of self-evolving mechanism.} 
Tab.~\ref{tab:abla_self_evolving} shows a comparison of our method against two key ablation baselines. The first baseline, GPT Internal Knowledge, relies solely on the GPT-5's pretrained knowledge without access to any external knowledge base. The second baseline, Hand-Crafted Prior, initializes the knowledge base with manually designed descriptions written by research experts to mimic GenArtist \cite{wang2024genartist}. Ours achieves gains of 0.11 over GPT Internal Knowledge and 0.03 over Hand-Crafted Prior, which validates that our self-evolving knowledge enables more evidence-based decision reasoning. 
% effectively overcoming the limitations of both the unstructured nature of pretrained priors and the inherent constraints of coverage and granularity in handcrafted rules.

\textbf{Effect of decision policy.} 
Shown in Tab.~\ref{tab:abla_router}, replacing the decision policy with a simple random tool selection policy leads to 0.23 relative performance drop on T2ICompBench++. It demonstrates that the knowledge-driven decision policy of OctoT2I is substantially more effective than an unguided multi-round trial and is the key driver of its leading performance.

\textbf{Effect of Exploration Space Pruning strategy.}
Tab.~\ref{tab:abla_pruning} shows exploration space pruning strategy reduces the average number of explored prompts per tool by 70.9\% and exploration time by 66.0\% compared to the ``w/o ESP" brute-force baseline, while achieving an identical overall score of 0.96. It proves our strategy enables a more efficient self-evolution that effectively discovers the capability frontiers.

% \textbf{Analysis of Performance-Efficiency Trade-off.}
% We analyze the two key hyperparameters controlling the performance-efficiency trade-off: the maximum number of rounds and the quality threshold ($\theta$). As shown in Fig.~\ref{fig:ablation_max_tries}, the GenEval score exhibits diminishing returns as the round limit increases, saturating after 8 rounds while the inference time grows steadily. Similarly, Fig.~\ref{fig:ablation_threshold} shows that increasing the quality threshold $\theta$ improves performance at the cost of higher latency, with the most significant performance gain occurring up to $\theta=0.8$. It is also worth noting that these parameters are adjustable, allowing users to flexibly control the trade-off based on their specific needs (\eg, prioritizing maximum quality vs. fastest generation).
\textbf{Analysis of Quality Threshold $\theta$.}
Fig.~\ref{fig:ablation_threshold} shows that increasing the quality threshold $\theta$ improves performance at the cost of higher latency, with the most significant performance gain occurring up to $\theta=0.8$. It is also worth noting that the parameter is adjustable, allowing users to flexibly control the trade-off based on their specific needs.

\vspace{-5pt}
\begin{table}[!t]
\centering
\caption{\textbf{Generalization performance} on the Wise benchmark.}
\vspace{-5pt}
\label{tab:wise_generalization}
\setlength{\tabcolsep}{2pt} % Adjust column spacing
\renewcommand{\arraystretch}{1.2}
\resizebox{\linewidth}{!}{
\begin{tabular}{l|cccccc|c}
\toprule
\textbf{Method} & \textbf{Cultural} & \textbf{Time} & \textbf{Space} & \textbf{Biology} & \textbf{Physics} & \textbf{Chemistry} & \textbf{Average}$\uparrow$ \\
\midrule
SD3.5-M & 0.43 & 0.50 & 0.52 & 0.41 & 0.53 & 0.33 & 0.45 \\
SD3.5-L & 0.44 & 0.50 & 0.58 & 0.44 & 0.52 & 0.31 & 0.46 \\
BAGEL & 0.44 & \best{0.55} & \best{0.68} & 0.44 & 0.60 & 0.39 & 0.52 \\
\rowcolor{blue!10}
\textbf{Ours} & \best{0.51} & \best{0.55} & 0.67 & \best{0.48} & \best{0.66} & \best{0.42} & \best{0.54} \\
\bottomrule
\end{tabular}
}
\vspace{-5pt}
\end{table}

\subsection{Generalization, Scalability, and Robustness}

\textbf{Generalization.}
Tab.~\ref{tab:wise_generalization} shows that OctoT2I achieves a competitive performance of 0.54, where the framework is directly adopted on the Wise benchmark. This demonstrates that OctoT2I is effective as a general-purpose framework that exhibits strong generalization capabilities.

\begin{table}[!t]
\centering
\caption{\textbf{Efficient New Tool Learning} on the Wise benchmark.}
\vspace{-5pt}
\label{tab:wise_new_tool_learning}
\setlength{\tabcolsep}{5pt} 
\renewcommand{\arraystretch}{1.2}
\scalebox{0.7}{
\begin{tabular}{lcc} 
\toprule
\textbf{Method} & \textbf{Average}$\uparrow$ & \textbf{$\Delta$ Explored Prompts} \\
\hline \hline
Ours  & 0.54 & 0 \\
Ours w/ 1 New Tool & 0.61 & 60 \\
\rowcolor{blue!10}
\textbf{Ours w/ 2 New Tools} & \best{0.71} & 130 \\
\bottomrule
\end{tabular}
}
\vspace{-5pt}
\end{table}

\textbf{Fast Learning and Integration of New Tools.} 
Shown in Tab.~\ref{tab:wise_new_tool_learning}, OctoT2I enables efficient, autonomous learning and seamless integration of new T2I tools—without requiring expert-designed handcrafted priors or costly training on new datasets, as demanded by prior methods~\cite{jia2024chatgen,wang2024genartist}. With just 60 exploration prompts, it incorporates Flux1.dev, lifting performance from 0.54 to 0.61. Adding gpt-image-1 further boosts the score to 0.71. This result demonstrates OctoT2I's exceptional scalability, transforming the traditionally high-cost, expert-dependent process of tool adaptation into a fully automated, low-overhead mechanism.

\begin{table}[!t]
\centering
\caption{\textbf{Robustness to other LLMs} on the Wise benchmark.}
\vspace{-5pt}
\label{tab:wise_robustness_llm}
\setlength{\tabcolsep}{2pt}
\renewcommand{\arraystretch}{1.2}
\resizebox{\linewidth}{!}{
\begin{tabular}{l|cccccc|c}
\toprule
\textbf{Method} & \textbf{Cultural} & \textbf{Time} & \textbf{Space} & \textbf{Biology} & \textbf{Physics} & \textbf{Chemistry} & \textbf{Average}$\uparrow$ \\
\hline
\hline
Ours & 0.51 & 0.55 & 0.67 & 0.48 & 0.66 & 0.42 & 0.54 \\
Ours w/ qwen & 0.50 & 0.57 & 0.64 & \best{0.53} & \best{0.70} & \best{0.48} & 0.55 \\
\rowcolor{blue!10}
\textbf{Ours w/ gpt5} & \best{0.61} & \best{0.60} & \best{0.68} & 0.52 & 0.68 & 0.46 & \best{0.60} \\
\bottomrule
\end{tabular}
}
\vspace{-14pt}
\end{table}

\textbf{Robustness to Different LLMs.}
% Tab.~\ref{tab:wise_robustness_llm} validates that OctoT2I does not depend on a specific LLM. We replace the default LLM in the decision module with Qwen3-VL-235B-A22B and GPT-5. The results indicate that after changing the LLM, the performance of OctoT2I improves from 0.54 to 0.55 and 0.60, respectively. This finding shows that the superior performance stems from the universality and effectiveness of our hierarchical knowledge and routing architecture, demonstrating the robustness of our framework.
Tab.~\ref{tab:wise_robustness_llm} demonstrates the LLM-agnostic robustness of OctoT2I. When the default LLM is replaced with Qwen3-VL-235B-A22B (w/ qwen) and GPT-5 (w/ gpt5), the performance improves to 0.55 and 0.60, respectively. This proves that the framework's performance gain is attributed to its universal hierarchical knowledge and routing architecture, rather than a specific LLM.

% \textbf{More details} about ablation, analysis of other hyperparameters, policy distillation and qualitative comparison are in \supp.

\begin{figure}[!t]
% \vspace{-5pt}
\centering
\includegraphics[width=0.9\linewidth]{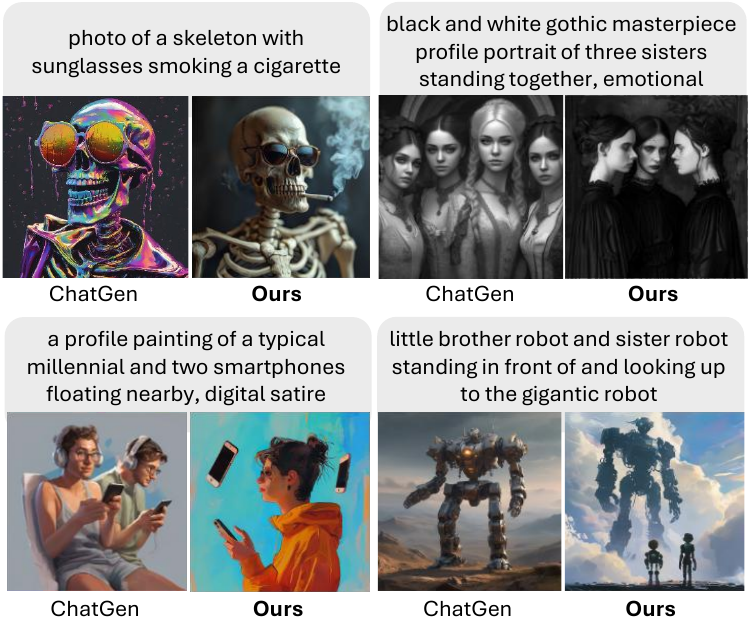}
\vspace{-5pt}
\caption{\textbf{Visual comparison} on real-world user prompts.}
\vspace{-5pt}
\label{fig:user_study}
\end{figure}

\begin{table}[!t]
\centering
\setlength{\tabcolsep}{2pt}
\caption{\textbf{User study results and efficiency comparison} of two agentic methods with real user prompts.}
\label{tab:user_study}
% \vspace{-8pt}
\resizebox{0.8\linewidth}{!}{
\begin{tabular}{l|ccc}
\toprule
\textbf{Method} & \textbf{Total Votes $\uparrow$} & \textbf{Voting Rate (\%)$\uparrow$} & \textbf{Avg. Time(s) $\downarrow$} \\
\hline \hline
ChatGen & 266 & 29.6 & 53.34  \\
\rowcolor{blue!10}
Ours  & \best{634} & \best{70.4} & \best{18.45} \\
\bottomrule
\end{tabular}}
\vspace{-14pt}
\end{table}

\subsection{User Study on Real-World User Prompts}
Tab.~\ref{tab:user_study} shows the result of a user study involving 30 researchers and 30 real-world user prompts. Our method achieves a 70.4\% voting rate, significantly outperforming ChatGen (29.6\%). The visual comparison in Fig.~\ref{fig:user_study} further demonstrates that OctoT2I produces better-aligned and more aesthetic visual results. Furthermore, OctoT2I (18.45s) is nearly $3\times$ faster than ChatGen (53.34s), proving effectiveness in co-optimizing performance and efficiency.

% \textbf{More details} regarding the hand-crafted prior baseline, more ablation study, user study setup and more visualization comparison are provided in the \supp.
% \textbf{Additional details} regarding the experimental setup, more ablation studies, discussions, and visualization comparisons are provided in the \supp.

\section{Conclusion}
This paper introduces \textbf{OctoT2I}, a novel agentic T2I framework that co-optimizes performance and efficiency, addressing key limitations of prior methods. We present two core innovations: (1) a \textbf{self-evolving mechanism} enabling autonomous knowledge acquisition via self-interaction, and (2) a \textbf{stateful, multi-round agentic router} that intelligently navigates the generation process using this knowledge. Experiments demonstrate that OctoT2I achieves leading performance while significantly reducing inference costs. Our work shows that self-evolving mechanisms and stateful routing represent a promising direction toward both more capable and efficient T2I systems.
As future work, we plan to explore extending this T2I framework to other generative domains, such as image editing and 3D generation.

% \begin{table*}[t]
% \centering
% \caption{\textbf{Efficiency comparison across different methods.} $\downarrow$ indicates higher is better.}
% \label{tab:efficiency}
% \setlength{\tabcolsep}{8pt}
% \renewcommand{\arraystretch}{1.2}
% \begin{tabular}{l|cccccccc}
% \toprule
% \textbf{Metrics} & \textbf{SDXL} & \textbf{BAGEL} & \textbf{Flow-GRPO} & \textbf{Idea2Img} & \textbf{GoT} & \textbf{ChatGen} & \textbf{GenArtist} & \cellcolor{blue!10}\textbf{Ours} \\
% \midrule
% Average Time (s)$\downarrow$ & \sbest{17.67} & 377.39 & 19.07 & 453.22 & 37.20 & 20.22 & 117.29 & \cellcolor{blue!10}\best{10.02} \\
% CO$_2$e (g)$\downarrow$      & \sbest{611.06} & 11198.25 & 878.72 & 12033.28 & 1258.47 & 783.60 & 4403.95 & \cellcolor{blue!10}\best{559.50} \\
% kWh $\cdot$ PUE$\downarrow$  & \sbest{1.41} & 25.86 & 2.02 & 27.79 & 2.91 & 1.81 & 10.17 & \cellcolor{blue!10}\best{1.29} \\
% \bottomrule
% \end{tabular}
% \end{table*}

{
    \small
    \bibliographystyle{ieeenat_fullname}
    \bibliography{main}
}
\end{document}